\documentclass[letterpaper, 10 pt, conference]{ieeeconf}

\IEEEoverridecommandlockouts                              %

\title{\LARGE \bf
Kidnapping Deep Learning-based Multirotors using Optimized Flying Adversarial Patches
}

\author{Pia Hanfeld$^{1,2}$, Khaled Wahba$^{2}$, Marina M.-C. Höhne$^{3,4}$, Michael Bussmann$^{1}$, and Wolfgang Hönig$^{2}$%
\thanks{$^{1}$ CASUS, Helmholtz-Zentrum Dresden-Rossendorf, $^{2}$ Technical University Berlin, $^{3}$ Leibniz Institute for Agricultural Engineering and Bioeconomy, $^{4}$ University of Potsdam}
\thanks{Corresponding author: p.hanfeld@hzdr.de}
\thanks{We thank Akmaral Moldagalieva for help with camera calibration.}%
\thanks{Code:\url{https://github.com/IMRCLab/flying_adversarial_patch/}}
\thanks{Video: \url{ https://youtu.be/Yj5rTcEJ0XE}}
\thanks{The research was partially funded by the Deutsche Forschungsgemeinschaft (DFG, German Research Foundation) - 448549715. Furthermore, it was partially funded by the Center for Advanced Systems Understanding (CASUS), financed by Germany’s Federal Ministry of Education and Research (BMBF), and by the Saxon state government out of the State budget approved by the Saxon State Parliament.
}%
}

\usepackage[bookmarks=true]{hyperref}
\usepackage{amssymb}
\usepackage{amsmath}
\usepackage{multirow}
\usepackage[detect-weight=true, detect-family=true,detect-inline-weight=math]{siunitx}

\usepackage[detect-weight=true, detect-family=true,detect-inline-weight=math]{siunitx}

\usepackage[ruled,vlined,linesnumbered]{algorithm2e}
\SetInd{0.1em}{0.4em}
\SetAlFnt{\footnotesize\sf}
\SetKwRepeat{Do}{do}{while}
\SetKw{KwStep}{step}
\SetKwInput{KwData}{Input}
\SetKwInput{KwResult}{Result}
\SetKwBlock{RepeatForever}{repeat}{}
\SetKw{Continue}{continue}
\SetKwComment{Comment}{$\triangleright$\ }{}
\SetCommentSty{itshape}

\usepackage[capitalise]{cleveref} %
\crefname{equation}{}{} %
\usepackage{flushend}

\usepackage[natbib=true,
    style=ieee,
    citestyle=numeric-comp,
    backend=bibtex,
    maxbibnames=10,
    maxcitenames=1,
    mincitenames=1,
    doi=false,
    url=false,
    giveninits=true]{biblatex}
\renewcommand*{\bibfont}{\footnotesize}

\usepackage{tikz}
\usetikzlibrary{fit,calc}
\newcommand{\vp}{\mathbf{p}}    %
\newcommand{\vtheta}{\boldsymbol{\theta}}
\newcommand{\vt}{\boldsymbol{t}}

\newcommand{\mC}{\mathbf{C}}    %
\newcommand{\mP}{\mathbf{P}}    %
\newcommand{\mT}{\mathbf{T}}    %
\newcommand{\mA}{\mathbf{A}}    %

\newcommand{\sC}{\mathcal{C}}   %
\newcommand{\sS}{\mathcal{S}}   %
\newcommand{\sE}{\mathcal{E}}   %
\newcommand{\sP}{\mathcal{P}}   %
\newcommand{\sT}{\mathcal{T}}   %

\newcommand{\vf}{\mathbf{f}}    %
\DeclareMathOperator*{\place}{place}
\DeclareMathOperator*{\ADAM}{ADAM}
\DeclareMathOperator*{\random}{random}

\DeclareMathOperator*{\argmin}{argmin}
\DeclareMathOperator*{\softmin}{softmin}

\DeclareMathOperator*{\FrontnetController}{ComputeFrontnetSetpoint}
\DeclareMathOperator*{\GetPatchPosition}{ComputePatchPose}
\DeclareMathOperator*{\RandomPerspective}{RandomPerspective}

\begin{document}

\maketitle
\thispagestyle{empty}
\pagestyle{empty}

\begin{abstract}

Autonomous flying robots, such as multirotors, often rely on deep learning models that make predictions based on a camera image, e.g. for pose estimation. These models can predict surprising results if applied to input images outside the training domain. This fault can be exploited by adversarial attacks, for example, by computing small images, so-called adversarial patches, that can be placed in the environment to manipulate the neural network's prediction. We introduce flying adversarial patches, where multiple images are mounted on at least one other flying robot and therefore can be placed anywhere in the field of view of a victim multirotor. By introducing the attacker robots, the system is extended to an adversarial multi-robot system. For an effective attack, we compare three methods that simultaneously optimize multiple adversarial patches and their position in the input image. We show that our methods scale well with the number of adversarial patches.
Moreover, we demonstrate physical flights with two robots, where we employ a novel attack policy that uses the computed adversarial patches to kidnap a robot that was supposed to follow a human.
\end{abstract}

\section{Introduction}
With the help of Deep learning (DL) models, unmanned aerial vehicles (UAVs) can detect and follow objects or human subjects in their environment. Tracking subjects is usually performed with estimations of the position and orientation angle, i.e., the \textit{pose}, of the subject. For pose estimation, the UAVs are equipped with a camera, and the camera images are then forwarded to a DL model predicting the pose relative to, e.g., the UAV's frame of reference. These poses can be utilized to generate new desired waypoints for the UAV, enabling the UAV to follow the subject.

It has been shown that DL models are susceptible to adversarial attacks~\cite{Szegedy2014, Goodfellow14}. Adversarial stickers or \textit{patches} have been widely applied in the autonomous vehicle domain to effectively force false predictions of neural networks (NNs) without the need to access the data or hardware~\cite{Brown2017, Sharma2022, Eykholt2018, Stein2023}. This can pose risks in particular when DL models are applied to control autonomous systems.
\begin{figure}
    \centering
    \includegraphics[width=\linewidth]{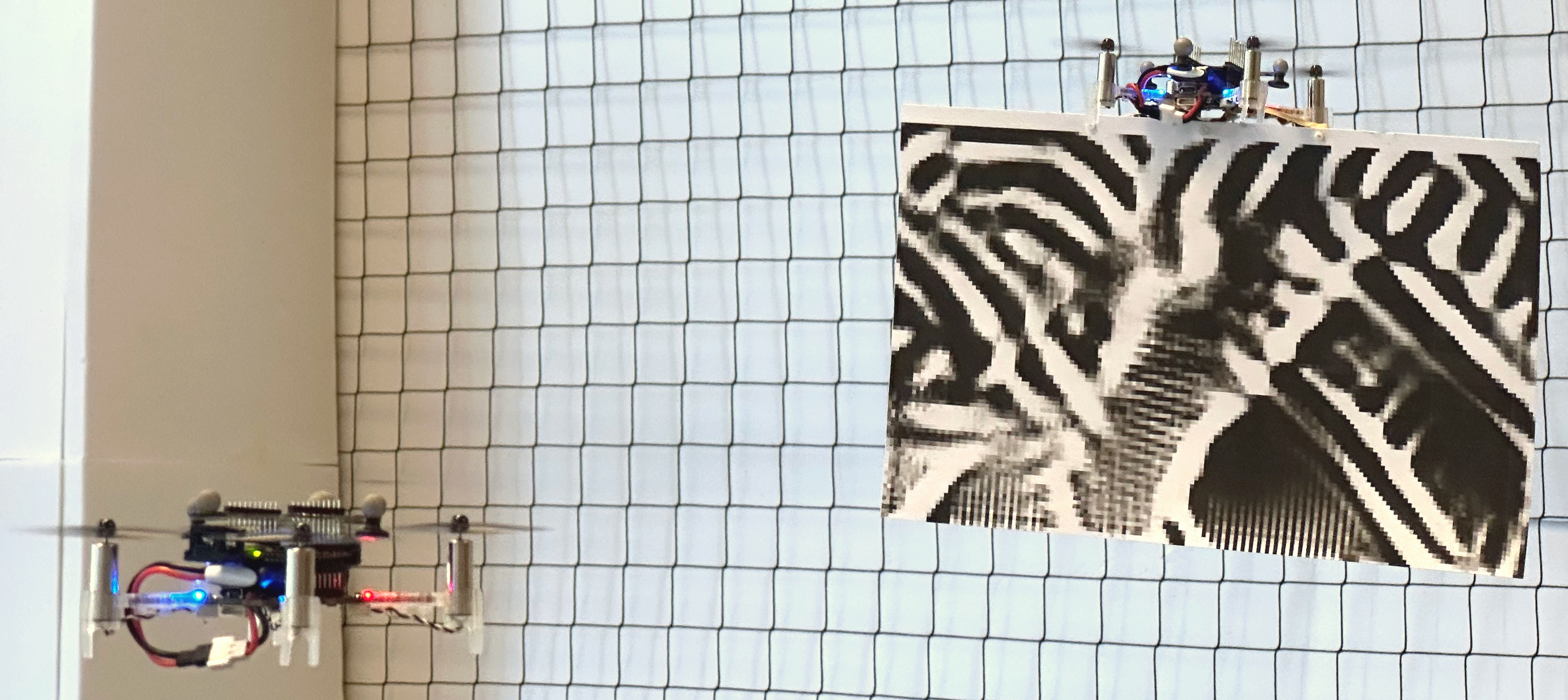}
    \caption{The attack scenario of this work: a multirotor carries two adversarial patches (printed one per side). The adversarial patches are able to prevent the NN from predicting the relative pose of a human subject in the input image, regardless of the human's position in the image. Instead, the NN predicts a pose previously chosen by the attacker policy that selects a patch and the relative position of it with respect to the victim.}
    \label{fig:overview}
\end{figure}

In this work, we propose to place adversarial patches in the field of view of a \textit{victim} multirotor's camera to force it onto a new desired trajectory, i.e., \textit{kidnapping} it, utilizing at least one additional \textit{attacker} multirotor. The computed adversarial patches can be printed and affixed to the attacker (see \cref{fig:overview}). Placing the patches in the environment in this way allows for a more flexible and prolonged attack scenario compared to the permanently placed patches of the state-of-the-art methods. Specifically, the attackers can directly control the placement of the patch by changing their relative positions~\cite{Hanfeld2023}. Additionally, we expect that multiple adversarial patches allow for more controllability over the victim. Switching between patches can be easily achieved by exploiting the yaw degree of freedom.

We assume that an attacker multirotor is limited to carry up to four patches (in which case they would be affixed on the sides of a cuboid), due to limited payload capabilities and to ensure that at most one of these patches is visible for the victim UAV.
If more than four patches are needed to achieve the desired controllability over the victim, multiple attacker multirotors can perform the attack.

In this paper, we focus on describing the optimization of multiple adversarial patches and their position in the input image and demonstrate with physical flights that the computed patches can be used to kidnap an autonomous drone.

The contributions of this work are as follows:
\begin{enumerate}
    \item We present novel methods to optimize multiple adversarial patches and their positions as a targeted adversarial attack.
    \item We apply our method to a publicly available, pre-trained regular and quantized NN for UAVs.
    \item We demonstrate our results in real-world experiments.
\end{enumerate}

\section{Related Work}
Adversarial attacks can be grouped into two categories: white-box attacks, when using the known network parameters, and black-box attacks, where the network's parameters are unknown.
Adversarial perturbations can be calculated to be applicable on a single input image only or on a whole image dataset, referred to as universal adversarial perturbation (UAP)~\cite{Moosavi-Dezfooli2017}.
Typically, classic (universal) adversarial perturbations are designed to be invisible to the human eye to achieve a stealthy attack. In contrast, adversarial patches are usually not hidden and appear like stickers or murals~\cite{Eykholt2018}.

Adversarial patches are special universal adversarial perturbations that can be printed and placed anywhere in the environment to be visible in the input images while maintaining a severe negative impact on the NN's performance. Adversarial patches are, therefore, a versatile tool to manipulate the prediction of a DL model without the need to alter the input image on a pixel level.

So far, many adversarial attacks focus on autonomous driving~\cite{Eykholt2018, Stein2023, Tu2020, Cao2022, Xu2020}. In the following, we focus on adversarial attacks for UAVs.

\citet{Raja2021} introduce a white- and black-box attack algorithm to an image classification NN deployed onboard of a UAV for bridge inspection. The NN's task is to identify regions prone to physical damage in the images. The output of the NN, therefore, does not influence the control of the UAV. The authors restrict the generated adversarial patches to be placed on physical objects only. They search within the whole image space for the optimal positions for the patches. %

\citet{Tian2021} introduce two white-box, single-image adversarial attacks on the publicly available DroNet~\cite{loquercio2018dronet}. Both methods can impact the steering angle and collision probability predicted by the NN. Their attacks force the network's predictions to differ as much as possible from the ground-truth predictions--they perform a so-called untargeted attack. In this case, the attacker has no control over the outcome of the attack and might even improve the performance of the NN. 

\citet{Nemcovsky2022} perform an adversarial patch attack on the Visual Odometry system used for navigation in simulation and real-world experiments. The patches were placed at fixed positions in both environments utilizing ArUco markers. In both examples, they were able to show that the attacks had a severe negative impact on the predicted traveled distance. Other than the previously described works, they take the influence on a sequence of images and different view angles into account.

Compared to current approaches, we introduce a targeted, white-box adversarial patch attack, that enables a potential real-world attacker to gain full control over a victim UAV. Instead of fixing one patch to be placed at a a-priori defined position, we optimize \emph{multiple patches} to be applicable to \emph{multiple positions} in the camera images. Furthermore, our method includes the optimization of these positions, as well as an effective attack policy.

\section{Background}

In the following, we introduce the NN we attack, PULP-Frontnet~\cite{Palossi2022}, a DL model that tracks a human and has been deployed on small multirotors. Moreover, we provide an introduction to adversarial patch placement.

\subsection{PULP-Frontnet}

PULP-Frontnet is a publicly available DL model implemented in PyTorch and developed for a nano multirotor--the Crazyflie by Bitcraze\footnote{\url{https://www.bitcraze.io/products/crazyflie-2-1/}}. The NN performs a human pose estimation task on the input images, predicting a human subject's 3D position and yaw angle in the UAV's frame. This prediction is later used to generate new setpoints for the UAV's controller which are subsequently used to follow the human subject. Therefore, the prediction of PULP-Frontnet directly influences the UAV's behavior.
For the generation of the adversarial patches of this work, we utilized the testset provided in~\cite{Palossi2022} together with our own data. The testset, consists of images of size $\mC_i\in\mathbb{R}^{160 \times 96}$ cropped at the same vertical position and without further augmentations, unlike the training dataset. These images are similar to a setting for a real-world attack. Our dataset, referred to as $\sC$, contains the 4028 images provided by~\cite{Palossi2022} and additional 2116 images collected in flight in our lab to increase the robustness of the optimized patches to a change of environment.
The forward pass of the NN $\vf_{\vtheta}(\cdot)$ can be defined as
\begin{align}
    \hat{\vp}^h &= \vf_{\vtheta}(\mC)\,,
\end{align}
where $\vtheta$ refer to the parameters of the NN, and the output is the estimated pose of the human subject $\hat{\vp}^h = (\hat{x}^h, \hat{y}^h, \hat{z}^h, \hat{\phi}^h)^T$ in the UAV's coordinate frame, where  $\hat{x}^h$ determines the distance in meters (depth), $\hat{y}^h$ the horizontal, $\hat{z}^h$ the vertical distance, and $\hat{\phi}^h$ the orientation angle around the $z$-axis of the subject to the UAV.
The predicted $\hat{\vp}^h$ is used to compute a setpoint that keeps the human in the center of the camera view. Thus, the UAV will follow the human subject depending on the predicted $\hat{\vp}^h$.
Therefore, manipulating the neural network output will directly influence the motions of the UAV.
\begin{figure*}
    \centering
    \includegraphics[width=0.9\textwidth]{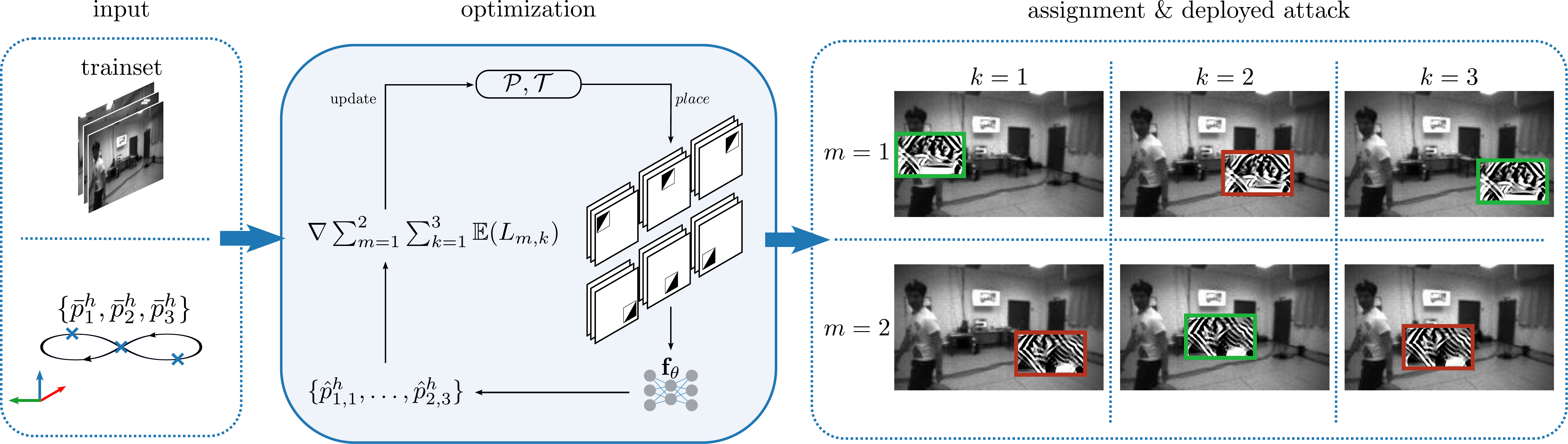}
    \caption{We compute i) a set of optimal adversarial patches $\sP^*$ and ii) a set of optimal transformation matrices $\sT^*$, such that an attacker can place all optimal patches $\sP^*$ in any image from the dataset $\sC$ and the error for the $k^{\text{th}}$ target position $\bar{\vp}^h_k$ and the predicted pose $\hat{\vp}^h_k$ under the attack is minimal. Left: The inputs are the trainset $\sS$ and the set of target positions $\{\bar{\vp}^h_1, \ldots, \bar{\vp}^h_K\}$, with $K=3$. Middle: $\sP$ and $\sT$ are optimized such that the expectation $\mathbb{E}$ of the loss $L_{m,k} = \|\bar{\vp}^h_k- \hat{\vp}^h_{m,k}\|_2$ becomes minimal for all $K$ target positions. The results are the optimal set of both patches $\sP^*$ and transformation matrices $\sT^*$. Right: During the optimization, we calculate the probability $\softmin(L_{1:M, k})$ for how likely a certain $\mP_m$ will achieve the current $\bar{\vp}^h_k$ implicitly while computing the expectation. This enables the optimal assignment $\rho: k \mapsto m$. The final placed patches in $\sP^*$ that are more likely to achieve the targets are marked with green rectangles, the patches that are more unlikely are marked with red rectangles.}%
    \label{fig:scetch_approach}
\end{figure*}
\subsection{Differentiable Adversarial Patch Placement}\label{sec:fixed_approach}
\citet{Thys2019} introduce an adversarial patch attack that prevents the detection of a human subject in camera images. They perform their attack on YOLOv2~\cite{redmon2017yolo9000}, a common NN for object detection. It predicts bounding boxes and classes for each object in an input image. Here, the main goal of the attack is to minimize the classification and a so-called objectness score for the subject. Minimizing the class score causes the model to misclassify the subject as another class. The objectness score indicates the likelihood of an object in the detected region.
Similarly to~\cite{Raja2021}, they restrict the patches to be placed on humans in the images. To improve robustness towards different view angles, lighting conditions, rotation, and scaling, they place and transform their patches with a method inspired by Spatial Transformers Networks~\cite{jaderberg2015spatial}.
The patches are placed and transformed with an affine transformation from a randomly generated transformation matrix. Additionally, they add random noise to counter the influence of noise added by the printer, the camera, and the lighting conditions. %

The affine transformation matrix is defined as \begin{align}
    \mT &= \begin{pmatrix}
        s \cos\alpha & -s\sin\alpha & t_x \\
        s\sin\alpha & s\cos\alpha & t_y \\
        0 & 0 & 1
    \end{pmatrix}\,,
\end{align}
where $s$ is a scale factor, $\alpha$ is the rotation angle, and $ \vt = (t_x,  t_y)^T$ is the translation vector.

Let $\mP\in\mathbb{R}^{w \times h}$ be a grayscale adversarial patch with width $w$ and height $h$. The patch can be placed in image $\mC$
\begin{align}
    \mC' &= \place(\mC, \mP, \mT)\,,
\end{align}
where $\place$ is the differentiable placement function and $\mC'$ is the resulting modified image. Note that the whole transformation process is differentiable.

We propose to additionally optimize the parameters of the transformation matrix, e.g., the scale, rotation, and translation vector instead of randomly choosing them during the training process. These parameters implicitly define an attack policy, thus optimizing them leads to a more effective and realistic attack scenario. Furthermore, we optimize a set of adversarial patches $\sP$ resulting in more control over the DL models predictions.

\section{Approach}\label{sec:approach}

The main objective of our method is to find i) a set of adversarial patches and ii) a policy for the attacker. The policy defines the attacker's behavior based on the current and the desired positions of the victim UAV with the goal to move it towards the desired position. This can be achieved by placing a selected patch with the best transformation matrix from a set of optimized matrices. %
The placed patch changes the NN output to the desired pose, such that the attacker gains full control over the victim UAV.

We set the rotation angle of all transformations $\mT$ to $\alpha = 0$, because rotation angles $\alpha \neq 0$ would require specifying a desired roll or pitch angle of the attacker UAV, which is not a controllable degree of freedom for a standard multirotor.

We split the dataset into two disjoint parts $\sC=\sS \cup \sE$ for training ($\sS$) and testing ($\sE$), respectively.

\subsection{Problem Definition}

Given the target prediction ($\bar{\vp}^h$) defined by the attacker and the current position of the human subject ($\vp^h$), our goal is to obtain i) $M$ patches $\sP$, and ii) a policy for the attacker that computes the desired pose of the attacking UAV, provided the current and desired positions of the victim: $\bar{\vp}^a = \pi^a(\vp^h, \bar{\vp}^h)$.
The objective is to minimize the tracking error of the victim between its current pose $\hat{\vp}^v$ and the desired pose $\bar{\vp}^v$, i.e., $\int_t \| \bar{\vp}^v(t) - \hat{\vp}^v(t)\| dt$.

For optimizing the patches, we manually specify $\pi^a$: the attacker UAV moves such that one of the patches it carries will be visible in the victim camera frame with a computed transformation $\mT$ (translation and scaling). Then the above can be simplified as follows.
Given are the $K$ target positions $\{\bar{\vp}^h_1, \ldots, \bar{\vp}^h_K\}$. Our goal is to compute i) a set of $M$ patches $\sP = \{\mP_1, \ldots, \mP_M\}$, ii) $M\times K$ transformations $\sT=\{\mT_{1,1}, \ldots, \mT_{M,K}\}$ (one for each patch and each target position), and iii) an assignment $\rho: k \mapsto m$ that decides which patch to use for each target position. %
For the assignment, we use a probabilistic formulation.
Thus, the objective is to minimize %
\begin{align}
    \label{eq:loss}
    L = & \frac{1}{|\sS|}\sum_{\mC\in\sS}\sum_{k=1}^K \mathbb{E}(L_k),
\end{align}
where $\mathbb{E}(L_k)$ is the expectation of the loss for the target position $k$.
We can approximate this expectation by using a softmin function
\begin{align}\label{eq:expectation}
    \mathbb{E}(L_k) = \sum_{m=1}^M \mathbb{P}(L_{m,k}) L_{m,k} = \softmin(L_{1:M,k}) \cdot L_{1:M,k},
\end{align}
where $L_{1:M,k}$ is the vector $[L_{1,k},\ldots,L_{M,k}]$ and $L_{m,k}$ is the prediction error when using patch $m$ with respect to the target position $k$ as defined in \cref{alg:split}, \cref{alg:Lmk}.

In the following, we introduce three optimization algorithms for calculating $\sP$ and $\sT$.%

\begin{algorithm}[t]
    \caption{Split Optimization} 
    \label{alg:split}
    \DontPrintSemicolon
     \KwData{$N$, $M$, $K$, $R$, $\sS$, $\sE$, $\vf_{\vtheta}$, $\{\bar{\vp}^h_1, \ldots, \bar{\vp}^h_K\}$ } %
     \KwResult{$\sP^*$, $\sT^*$}

    $A_{m,k} \gets \text{True}\,, \forall m \in \{1, \ldots, M\},\; \forall k \in \{1, \ldots, K\} $\label{alg:initial_assignment}\\

    \ForEach{$n \in \{1, \ldots, N\}$}{\label{alg:split:P_only_start}
        \Comment{Optimize Patch}
        $L \gets 0$\\

    \ForEach{$\mC \in \sS$}{
        $E \gets 0, a \gets 0$\\
        \ForEach{$m, k \text{ where } A_{m,k} = \text{True}$}{
            $L_{m,k} \gets \|\bar{\vp}^h_k-\vf_{\vtheta}(\place(\mC, \mP_m, \mT_{m,k}))\|_2$ \label{alg:Lmk}

                $E \gets E + L_{m,k} \cdot e^{-L_{m,k}}$\Comment*{$\softmin(\cdot)$ numerator}
                $a \gets a + e^{-L_{m,k}}$\Comment*{$\softmin(\cdot)$ denominator}
         }
         \If{$a \neq 0$}{
            $L \gets L + \frac{E}{a}$\Comment*{sum expectation for all $\mC$}
         }
        }

        $\sP \gets \ADAM(\nabla_{\sP} \frac{L}{|\sS|})$\label{alg:split:P_only_end}\Comment*{update $\sP$}

        \Comment{Optimize Positions}
        \ForEach{$m, k \text{ where } A_{m,k} = \text{True}$}{\label{alg:split:T_begin}
            \ForEach{$r\in\{1, \ldots, R\}$}{\label{alg:split:restarts}
            $\mT \gets \begin{cases}
                \mT_{m,k} & \text{if } r = 1\\
                \random\mT() & \text{else}
            \end{cases}$
                \label{alg:split:restarts_end}

            $L_r \gets \frac{1}{|\sS|}\sum_{\mC\in\sS}\|\bar{\vp}^h_k-\vf_{\vtheta}(\place(\mC, \mP_m, \mT))\|_2$\label{alg:split:loss_restart}\\ \Comment*{loss for current $\mT$}
            $\mT_r \gets \ADAM(\nabla_{\mT}L_r)$\label{alg:split:update_T} \Comment*{update $\mT$}
            }
          $\mT_{m,k} \gets \argmin_{\mT_r} \{L_1, \ldots, L_R\}$\label{alg:split:update_mTk} \Comment*{choose best $\mT_r$}
         }

        \Comment{Optimal Assignment}
        $C_{m,k} \gets \frac{1}{|\sE|}\sum_{\mC\in\sE}\|\bar{\vp}^h_k-\vf_{\vtheta}(\place(\mC, \mP_m, \mT_{m,k}))\|_2,\forall m, k$\label{alg:testloss}\\

        $A_{m,k} \gets \begin{cases}
            \text{True} & (m' \sim \softmin(C_{1:M,k})) \land m' = m \\
            \text{False} & \text{else}
        \end{cases}$\label{alg:update_assignment}

    }
\Return $\sP^*$, $\sT^*$
\end{algorithm}

\subsection{Joint Optimization}\label{sec:joint}

One approach is the simultaneous optimization of the set of adversarial patches and the corresponding transformation matrices. %
Hence, $\sP$ and $\sT$ are updated after each iteration using the gradient of the loss $L$, see \cref{eq:loss}, with respect to $\sP$ and $\sT$: 
\begin{align}
    \sP, \sT \gets \ADAM(\nabla_{\sP, \sT} L).
\end{align}

Note that the parameters $\vtheta$ of the NN $\vf$ are fixed and not changed during the optimization. The optimization workflow is displayed in \cref{fig:scetch_approach}.
To increase the numerical robustness of $\sP$ and $\sT$, we add small random perturbations to each image after the patch placement and to each $\mT_{m,k}$, using a normal distribution.
In \cref{eq:expectation}, we compute the probability $\mathbb{P}(L_{m,k})\,, \forall m,k$ for how likely a certain $\mP_m$ placed at $\mT_{m,k}$ will achieve the current target $\bar{\vp}^h_k$. Choosing the optimal patch $\mP^*_{m^*}$ for each target $k$ from the set of all optimal patches $\sP^*$ can be selected such that $m^*= \argmin_{m} L_{1:M,k}$, allowing for the optimal assignment $\rho: k \mapsto m$.

\subsection{Split Optimization}
The joint optimization might lead to a local optimum instead to the global solution for the optimal position of the adversarial patches due to the local nature of the gradient-based optimization and the highly non-convex function $\vf_{\vtheta}$. Therefore, we propose to split the training algorithm into three parts: We first only optimize the adversarial patches $\sP$ for a fixed set of transformations $\sT$. For a user-defined number of iterations $N$, the patches are optimized for the $K$ desired positions $\{\bar{\vp}^h_1, \ldots, \bar{\vp}^h_K\}$, while the transformation matrices $\sT$ are fixed, see \Cref{alg:split}, \cref{alg:split:P_only_start} - \cref{alg:split:P_only_end}. We initialize $\mA$ by assigning all $M$ patches in $\sP$ to all $K$ targets in \cref{alg:initial_assignment}. 

Inspired by the coordinate descent algorithm~\cite{Wright2015}, we perform \textit{random restarts} as a second step (from \cref{alg:split:T_begin} to \cref{alg:split:update_mTk}). Not only are the current parameters of the matrices in $\sT$ optimized but, for $R-1$ restarts, chosen randomly and trained for an iteration over $\sS$ (\cref{alg:split:restarts} - \cref{alg:split:update_T}). We then choose the transformation matrix $\mT_{m,k}$ that produces the minimal loss over all restarts $R$ (\cref{alg:split:update_mTk}). These random restarts aim to overcome local minima while searching for effective transformation matrices.

As a third step, we randomly assign a patch $m$ to a target $k$ by sampling $m$ from the multinomial probability distribution of the cost for the testset $\sE$ (\cref{alg:update_assignment}).
This finds better combinations of patches and their placement after the random restarts.  %

\begin{algorithm}[t]
    \caption{Attacker Policy} 
    \label{alg:policy}
    \DontPrintSemicolon
    \KwData{$\sT, \mA, \hat{\mT}_a, \hat{\mT}_v, \bar{\mT}_v$}
    \KwResult{$\bar{\mT}_a$}

    \ForEach{$m, k \text{ where } A_{m, k} = \text{True} $\label{alg2:start_frontnet_setpoints}}{
    $\mT_v(m,k) \gets \FrontnetController(\hat{\mT}_v, \bar{\vp}^h_k)$ \\ \label{alg2:end_frontnet_setpoints}
    }
    $m^*, k^* \gets \argmin_{m, k} \|\vp(\bar{\mT}_v) - \vp(\mT_v(m,k))\|_2$\label{alg2:tracking_error} \\

    $\mT_p \gets \GetPatchPosition(\mT_{m^*, k^*}, m^*)$\label{alg2:patchpos}\\

    $\bar{\mT}_a \gets \hat{\mT}_v \mT_p$\label{alg2:attackerpose}\\

    \Return $\bar{\mT}_a$
\end{algorithm}

\subsection{Hybrid Optimization}
The hybrid version of our method combines both the joint and split optimization. The patches and the transformation matrices are trained jointly for a fixed amount of iterations over $\sS$. The parameters of the transformation matrices $\mT_{m,k}$ are then fine-tuned while the patches in $\sP$ are fixed, analogous to the optimization described in \Cref{alg:split} starting from \cref{alg:split:T_begin}. The optimal assignment is identical to the one in the split optimization.

\subsection{Attacker Policy}
The attacker policy depends on the set of optimized transformation matrices $\sT$, the binary matrix $\mA$ representing the assignment $\rho$, the current pose of the attacker UAV $\hat{\mT}_a$, the current pose of the victim UAV $\hat{\mT}_v$, and the desired pose of the victim UAV $\bar{\mT}_v$ (all poses are in world coordinates). The goal is to calculate the desired pose of the attacker $\bar{\mT}_a$, such that the victim moves to its desired pose.

The function $\FrontnetController(\cdot)$ (see \Cref{alg:policy}, \cref{alg2:start_frontnet_setpoints} and \cref{alg2:end_frontnet_setpoints}) computes the setpoint the controller of the victim UAV will output, assuming our adversarial patches would achieve a loss of $0$ and Frontnet predicts exactly the specified targets $\bar{\vp}^h_k$ for all $m, k$, where $A_{m,k} =$ True.
We then compute the optimal $m^*$ and $k^*$ that minimize the tracking error between the current and desired victim pose (\cref{alg2:tracking_error}).
In $\GetPatchPosition(\cdot)$ (\cref{alg2:patchpos}), we compute the pose of the attacker relative to the victim multirotor with the corresponding $\mT_{m^*, k^*}$  utilizing the camera intrinsics, extrinsics, and distortion coefficients.
This function also takes the patch as input ($m^*$) and adjusts the orientation of the attacker such that the correct patch is visible in the victim's camera.
The attacker pose in world coordinates is then calculated from the matrix multiplication of $\hat{\mT}_v$ and $\mT_p$ (\cref{alg2:attackerpose}).

\begin{figure}
    \centering
    \includegraphics[page=18, width=0.23\textwidth, trim= 2 0 2 0, clip]{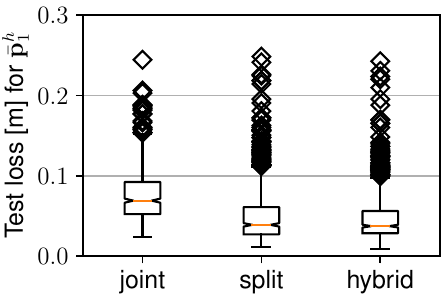}
    \includegraphics[page=19, width=0.23\textwidth, trim= 2 0 2 0, clip]{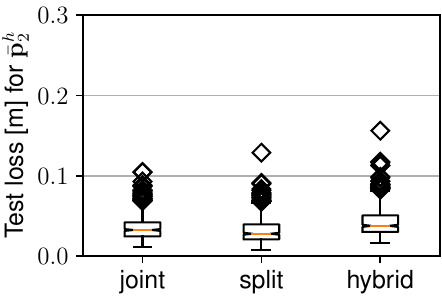}
    \caption{Resulting mean test loss for the proposed joint, split, and hybrid optimization. The joint optimization calculates less effective $\sP$ and $\sT$ for $\bar{\vp}^h_1$ compared to the split and hybrid optimization, resulting in a higher mean test loss for the joint optimization. The hybrid optimization outperforms the split optimization over both targets slightly.}
    \label{fig:exp1:all}
\end{figure}
\begin{figure}
    \centering
     \includegraphics[page=17, width=0.38\textwidth]{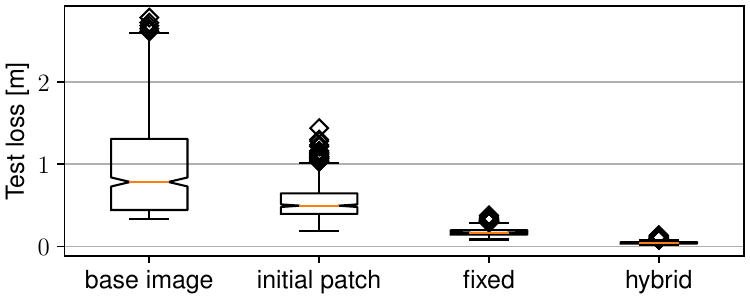}
     \caption{Resulting mean test loss for all $\mC\in\sE$ (i.e. \textit{base images} showing no patch) per target. For comparison, the \textit{initial patches} (i.e. patches showing only random noise) are placed with the optimal $\sT$ calculated with the hybrid approach, the algorithm that achieves the lowest mean test loss over all targets (see \cref{fig:exp1:all}). Optimizing the adversarial patches $\sP$ with the fixed approach leads to a less effective attack compared to the other optimization approaches.}
     \label{fig:exp1:basevsstartvshybrid}
\end{figure}

\section{Results}
We now analyze all of our proposed approaches, described in \cref{sec:approach}, and compare them to the method introduced in~\cite{Thys2019}, referred to as \textit{fixed optimization}.

We implement our method in PyTorch, utilizing the Adam optimizer~\cite{Kingma2014} to perform the gradient-based optimization. We compute all optimization approaches in parallel on compute nodes with access to 4 cores of an Intel 12-Core Xeon (3.0 GHz) CPU and a NVIDIA Tesla P100 GPU.

The learning rate for all approaches is set to $0.001$, regardless of which combination of parameters is optimized. %
The trainset $\sS$ contains $90\%$ of the images from the dataset $\sC$. Batches of size 32 are drawn uniformly at random from $\sS$ in each training step, and we train on all batches for one epoch. 

The training process of $K=2$ patches is repeated ten times with random initial seeds, referred to as \textit{trials}. For each approach, $\sP$ and $\sT$ are trained for $N=100$ iterations on the trainset $\sS$. Before placing the patches in $\sP$, we add random Gaussian noise with mean of $0$ and a standard deviation of $0.1$ to all matrices in $\sT$. To simulate the aerodynamic effects on the patches in flight, we add random perspective transformations to the transformed patches after they are placed with $\sT$. For the random perspective transformations, we rely on the PyTorch function $\RandomPerspective$ with a distortion scale of $0.2$ and a probability of $0.9$. Additionally, we add Gaussian noise to the manipulated images $\mC'$ with a standard deviation of $10$.

To ensure that the patches are visible in $\mC'$, we restrict the parameters of the transformation matrices in $\sT$. The scale factor $s$ is kept in $[0.2, 0.4]$ and the translation vector $\vt$ is kept in $(-1, 1)$.
We perform $R=20$ random restarts for the split and hybrid optimization. We validate the negative impact on the NN $\vf_{\vtheta}$ with the mean of the loss defined in \cref{eq:loss} on the unseen testset $\sE$ over the ten trials. The testset $\sE$ contains $10\%$ of the images in $\sC$.
We provide a supplemental video for our real-world experiment.

\subsection{Comparison of the Different Approaches}\label{sec:comparison}

We compare the proposed joint, split, and hybrid optimization introduced in \cref{sec:approach} to the fixed optimization. In the latter, only the adversarial patches $\sP$ are optimized--the transformation matrices are initialized randomly and fixed for all training steps. We use $K=2$ and the two target poses are $\bar{\vp}^h_1 = (1, -1, 0)^T$ and $\bar{\vp}^h_2 = (1, 1, 0)^T$. 

\begin{figure}
    \centering
    \includegraphics[width=0.45\textwidth]{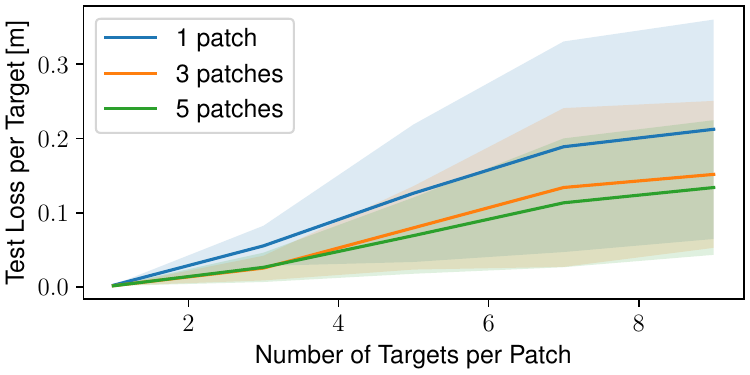}
    \caption{Mean test loss per number of targets over five trials. The shaded area depicts the variance of the test loss across the targets and trials. The attack is significantly effective for $1\leq K \leq 4$. The variance increases as soon as the target poses $\bar{\vp}^h_k$ are included, which are underrepresented in the trainset. The loss per target and the variance decreases with an increasing number of optimized patches. The difference between the losses for $M=3$ and $M=5$ patches is not as significant as the difference between $M=1$ and $M=3$.}
    \label{fig:exp2}
\end{figure}

\begin{table*}
    \centering
    \caption{Validation of different training approaches on the quantized NN. Methods marked with (Q) were trained on the quantized NN, and methods marked with (FP) were trained on the full-precision NN. Lower is better.} 
\begin{tabular}{l|c|c||c|c||c|c||c|c}
               & Fixed (FP) & Fixed (Q)      & Joint (FP)     & Joint (Q)      & Split (FP) & Split (Q)      & Hybrid (FP) & Hybrid (Q)     \\ \hline
mean test loss & 0.401      & \textbf{0.162} & 0.052          & \textbf{0.046} & 0.192      & \textbf{0.048} & 0.053       & \textbf{0.051} \\ \hline
mean std       & 0.286      & \textbf{0.066} & \textbf{0.004} & 0.024          & 0.138      & \textbf{0.012} & 0.039       & \textbf{0.015}
\end{tabular}
    \label{tab:quantized}
\end{table*}

The computed $\sP^*$ and $\sT^*$ utilizing the hybrid optimization on average leads to a $14\%$ lower loss compared to the joint optimization and a $0.5\%$ lower loss compared to the split optimization on the testset and is, therefore, the best approach for the described setup. The mean test loss over 10 trials for $\bar{\vp}^h_1$ and $\bar{\vp}^h_2$ is displayed in \cref{fig:exp1:all}. The fixed optimization computes $\sP^*$ that are less effective on all images from the testset compared to the other approaches, see \cref{fig:exp1:basevsstartvshybrid}. %

Due to the high number of random restarts $R$ performed during the split and hybrid optimization, the computation time on our hardware setup for 100 iterations over $\sS$ roughly takes 10 times longer than the joint optimization which is therefore the preferred method.

\subsection{Ablation Study}
We investigate the scalability of our approach with respect to the number of targets $K$ and the number of patches $M$, see \cref{fig:exp2}. The first nine targets are set to $\bar{\vp}^h_k = (1, y, z)^T$, where $y, z \in \{-1, 0, 1\}$, representing predictions in the corners and the center of the image $\mC$. The last target is set to $\bar{\vp}^h_{10} = (2, 0, 0)$, representing a prediction in the center of the image but further away from the victim UAV. The mean test loss per target decreases significantly if three patches (orange) instead of a single patch are utilized (blue). The loss decreases even further if $M = 5$ but the difference to $M = 3$ is not as significant. This demonstrates that multiple patches increase the control over the DL model's predictions and that even a single attacker multirotor is sufficient in most cases.

The variance increases with the number of targets because not all target positions can be equally easily achieved. The loss increases as soon as target poses $\bar{\vp}^h_k$ are included, which are underrepresented in the original trainset of PULP-Frontnet--especially $z > 0.5$ and $z < -0.5$. Additionally, $y < 0$ is more challenging for the adversarial patch attack since the human subjects included in the images of $\sC$ are primarily located at $y > 0$. %

\subsection{Effect on the Quantized NN}\label{sec:quantized}
Since the quantized version of the network will be deployed on the hardware (for details about quantization see \cite{Palossi2022}), the calculated patches might not have the same effect if optimized for the full-precision parameters of the NN due to quantization errors.
Therefore, the stored integer values of the provided quantized NN are loaded into a floating point PyTorch model of the same architecture as the previously attacked Frontnet, now referred to as $\vf_{\vtheta_q}$, and the same experiment as in \cref{sec:comparison} is repeated but for a single patch only, initialized showing a face from $\sC$. %
We now compare the influence of the patches calculated on $\vf_{\vtheta}$ with the influence of the patches calculated on $\vf_{\vtheta_q}$ on the quantized NN. In \cref{tab:quantized}, it can be observed that the patches calculated on $\vf_{\vtheta_q}$ indeed have a lower loss on the predictions of $\vf_{\vtheta_q}$. Here, the joint approach outperforms the other introduced methods. It produces a $12\%$ more effective patch compared to the patch computed for the full-precision NN. 

\subsection{Real-world experiments}
For the real-world experiments, we trained $M=2$ patches for $K=5$ targets. The desired trajectory for the victim is to move along a smooth curve in $y \in [-0.5, 0.5]$ \si{m}, while $x$ and $z$ are fixed. The overall goal is to force the victim UAV to follow the desired trajectory instead of performing its original task, i.e., following a human subject.

We validate the behavior of the victim UAV in three different experiments:
i) only a human and the victim UAV, ii) only the attacker and victim UAV, and iii) a human, the attacker, and the victim UAV are in the flight space.

For the first experiment, we validate that our implementation of PULP-Frontnet\footnote{
The pretrained NN is available but the firmware implementation is not.
} is working, i.e., the victim UAV is successfully following a human subject. While following the human, the mean $L_2$-error to the desired trajectory is \SI{0.57}{m} (std \SI{0.25}{m}). In the second experiment, we show that the victim follows the attacker UAV. The attacker is able to force the victim UAV onto the desired trajectory with a mean $L_2$-error of \SI{0.47}{m} (std \SI{0.24}{m}). For the third experiment, the mean $L_2$-error is \SI{0.34}{m} (std \SI{0.22}{m}). As observable in the supplemental video, the human is able to distract the victim UAV from following the attacker in the second half of the experiment. Due to the trajectory the human is following in that part, we were able to achieve a lower error to the desired trajectory for the victim UAV.
The tracking errors are high overall, because we are unable to move the attacker at high speeds since the attached patch causes significant drag.
Still, the presented patch was able to distract the victim UAV from tracking the human.

\section{Conclusion}
This paper introduces new methods for optimizing multiple adversarial patches and their positions simultaneously, a policy for an attacker UAV, and a demonstration in the real world.
Our empirical studies show that the described hybrid optimization approach outperforms the joint and split optimization for optimizing the patches and their positions. All introduced approaches exceed the state-of-the-art method, i.e., fixed optimization, for the described attack scenario. %
The optimized patches presented to the victim UAV at the optimized positions are able to successfully \textit{kidnap} the victim, i.e., force it to follow the desired target trajectory, even when a person is visible in the camera images.

In this work, we only utilize a single attacker UAV. In the future, the attacker policy can be enhanced to present multiple patches with multiple UAVs, creating a team of attacker UAVs. %

\printbibliography

\end{document}